# Graphical Condition for Identification in Recursive SEM

Carlos Brito and Judea Pearl

## Abstract

The paper concerns the problem of predicting the effect of actions or interventions on a system from a combination of (i) statistical data on a set of observed variables, and (ii) qualitative causal knowledge encoded in the form of a directed acyclic graph (DAG). The DAG represents a set of linear equations called Structural Equations Model (SEM), whose coefficients are parameters representing direct causal effects. Reliable quantitative conclusions can only be obtained from the model if the causal effects are uniquely determined by the data. That is, if there exists a unique parameterization for the model that makes it compatible with the data. If this is the case, the model is called identified. The main result of the paper is a general sufficient condition for identification of recursive SEM models.

## 1 Introduction

Structural Equation Models (SEM) is one of the most important tools for causal analysis in the social and behavioral sciences [2, 5, 8, 1, 6, 7]. Although most developments in SEM have been done by scientists in these areas, the theoretical aspects of the model provide interesting problems that can benefit from techniques developed in computer science.

In a structural equation model, the relationships among a set of observed variables are expressed by linear equations. Each equation describes the dependence of one variable in terms of the others, and contains a stochastic error term accounting for the influence of unobserved factors.

An attractive characteristic of SEM models is their simple causal interpretation. Specifically, the linear equation $Y = \beta X + e$ encodes two distinct assumptions: (1) the possible existence of (direct) causal influence of $X$ on $Y$; and, (2) the absence of (direct) causal influence on $Y$ of any variable that does not appear on the right-hand side of the equation. The parameter $\beta$ quantifies the (direct) causal effect of $X$ on $Y$. That is, the equation claims that a unit increase in $X$ would result in $\beta$ units increase of $Y$, assuming that everything else remains the same.

Let us consider an example taken from [10]. This model investigates the relations between smoking $(X)$ and lung cancer $(Y)$, taking into account the amount of tar $(Z)$ deposited in a person's lungs, allowing for unobserved factors to affect both smoking $(X)$ and cancer $(Y)$:

$$X = e_1$$
$$Z = aX + e_2$$
$$Y = bZ + e_3$$
$$cov(e_1, e_2) = cov(e_2, e_3) = 0$$
$$cov(e_1, e_3) = \gamma$$

The first three equations claim, respectively, that the level of smoking of a person depends only on factors not included in the model, the amount of tar deposited in the lungs depends on the level of smoking as well as external factors, and the level of cancer depends on the amount of tar in the lungs and external factors. The remaining equations say that the external factors that cause tar to be accumulated in the lungs are independent of the external factors that affect the other variables, but the external factors that have influence on smoking and cancer may be correlated.

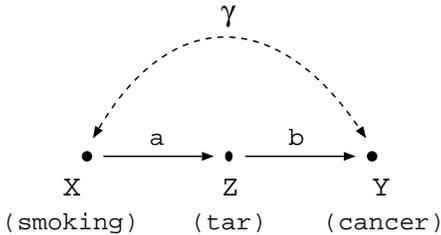

Figure 1: Smoking and lung cancer example

All the information contained in the equations can be expressed by a graphical representation, called *causal diagram*, as illustrated in Figure 1.

The process of data analysis using Structural Equation Models consists of four steps [7]: (1) specification of the model, (2) analysis of identification, (3) estimation of parameters, and (4) evaluation of fit. In this work, we will concentrate on the problem of Identification. The identification of a model is important because, in general, no reliable quantitative conclusion can be derived from a non-identified model.

## 1.1 Related Work

The question of identification has been the object of extensive research [6, 5, 10, 8, 11]. Despite all this effort, the problem still remains open. That is, we do not have a necessary and sufficient condition for identification in SEM.

Traditional approaches to the Identification problem are based on algebraic manipulation of the equations defining the model. Powerful algebraic methods have been developed for testing whether a specific parameter, or a specific equation in the model is identifiable [6, 9]. However, those methods are limited in scope. The rank and order criteria [6], for example, do not exploit restrictions on the error covariances (if such are available). Identification methods based on block recursive models [6, 11], for another example, insist on uncorrelated errors between any pair of ordered blocks.

Recently, some advances have been achieved on graphical conditions for identification [10, 4]. Examples of such conditions are the "back-door" and "single-door" criteria [10]. A problem with such conditions is that they are applicable only in sparse models, that is, models rich in conditional independence. The same holds for criteria based on instrumental variables (IV) [12], since these require search for variables (called *instruments*) that are uncorrelated with the error terms in specific equations.

## 1.2 Overview of Results

In our approach to the problem, we state Identification as an intrinsic property of the model, depending only on its structural assumptions. Since all such assumptions are captured in the graphical representation of the model, we can apply graph theoretic techniques to study the problem of Identification in SEM. Thus, our main result consist of a graphical condition for identification, to be applied on the causal diagram of the model.

The basic tool used in the analysis is Wright's decomposition of correlations, which allows us to express correlation coefficients as polynomials on the parameters of the model.

Based on the observation that these polynomials are linear on specific subsets of parameters, we reduce the problem of Identification to the analysis of systems of linear equations. As one should expect, conditions for linear independence of those systems (which imply a unique solution and thus identification of the parameters), translate into graphical conditions on the paths of the causal diagram.

## 2 Background

### 2.1 Structural Equation Models and Identification

A structural equation model $M$ for a vector of observed variables $\mathbf{Y} = [Y_1, \ldots, Y_n]'$ is defined by a set of linear equations of the form

$$Y_j = \sum_i c_{ji} Y_i + e_j \qquad , \text{ for } j = 1, \ldots, n.$$

Or, in matrix form, $\mathbf{Y} = C \cdot \mathbf{Y} + \varepsilon$, where $C = [c_{ji}]$ and $\varepsilon = [e_1, \ldots, e_n]'$.

The term $e_j$ in each equation corresponds to a stochastic error, assumed to have normal distribution with zero mean. The model also specifies independence assumptions for those error terms, by the indication of which entries in the matrix $\Psi = [\psi_{ij}] = Cov(e_i, e_j)$ have value zero.

In this work, we consider only recursive models, which are characterized by the fact that the matrix $C$ is lower triangular. This assumption is reasonable in many domains, since it basically forbids feedback causation.

The set of parameters of model $M$, denoted by $\Theta$, is composed by the (possibly) non-zero entries of matrices $C$ and $\Psi$.

A parameterization $\pi$ for model $M$ is a function $\pi : \Theta \to \Re$ that assigns a real value to each parameter of the model. The pair $\langle M, \pi \rangle$ determines a unique covariance matrix over the observed variables, given by [2]:

$$\Sigma_M(\pi) = \left(I - C(\pi)\right)^{-1} \Psi(\pi) \left[\left(I - C(\pi)\right)^{-1}\right]^T$$

where $C(\pi)$ and $\Psi(\pi)$ are obtained by replacing each non-zero entry of $C$ and $\Psi$ by the respective value assigned by $\pi$.

Now, we are ready to define formally the problem of Identification in SEM.

**Definition 1 (Model Identification)** *A structural equation model $M$ is* identified *if, for almost every parameterization $\pi$ for $M$, the following condition holds:*

$$\Sigma_M(\pi) = \Sigma_M(\pi') \implies \pi = \pi' \quad (1)$$

*More precisely, if we view parameterization $\pi$ as a point in $\Re^{|\Theta|}$, then the set of points in which condition (1) does not hold has Lebesgue measure zero.*

In general, if a model $M$ is non-identified, for each parameterization $\pi$ there exists an infinite number of distinct parameterizations $\pi'$ such that $\Sigma_M(\pi) = \Sigma_M(\pi')$. However, there are models in which, for almost every parameterization, there exist a finite number of distinct parameterizations that generate the same covariance matrix. According to the definition above, those models are classified as non-identified. However, it is important to distinguish this situation from the general case of non-identification. This motivates the following definition:

**Definition 2 (K-Identification)** *A structural equation model $M$ is* k-identified *if, for almost every parameterization $\pi$ for $M$, the number of distinct parameterizations that generate the covariance matrix $\Sigma_M(\pi)$ is at most $k$.*

$Z = e_1$
$W = e_2$
$X = aZ + e_3$
$Y = bW + cX + e_4$
$Cov(e_1, e_2) = \alpha \neq 0$
$Cov(e_2, e_3) = \beta \neq 0$
$Cov(e_3, e_4) = \gamma \neq 0$

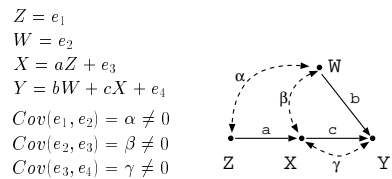

Figure 2: A causal diagram

## 2.2 Graphical Representation

The causal diagram of a model $M$ consists of a directed graph whose nodes correspond to the observed variables $Y_1, \ldots, Y_n$ in the model. A directed edge from $Y_i$ to $Y_j$ indicates that $Y_i$ appears on the right-hand side of the equation for $Y_j$ with a non-zero coefficient. A bidirected arc between $Y_i$ and $Y_j$ indicates that the corresponding error terms, $e_i$ and $e_j$, have non-zero correlation. The graphical representation can be completed by labeling the edges with the parameters of the model. Figure 2 shows a simple causal diagram.

A *path* between variables $X$ and $Y$ in a causal diagram consists of a sequence of edges $\langle e_1, e_2, \ldots, e_n \rangle$ such that $e_1$ is incident to $X$, $e_n$ is incident to $Y$, and every pair of consecutive edges in the sequence has a common variable. We say that the path points to $X$ if the edge $e_1$ has an arrow head pointing to $X$.

A path $p = \langle e_1, \ldots, e_n \rangle$ between $X$ and $Y$ is *valid* if variable $X$ only appears in $e_1$, variable $Y$ only appears in $e_n$, and every intermediate variable appears only once in the path.

The special case of a path composed only by directed edges, all of which oriented in the same direction, is called a *chain*.

We make use of a few family terms to refer to variables in particular topological relationships. Specifically, if the edge $X \to Y$ is present in the causal diagram, then we say that $X$ is a *parent* of $Y$. Similarly, if there exists a chain from $X$ to $Y$, then $X$ is an *ancestor* of $Y$, and $Y$ is a *descendant* of $X$.

Given a path $p$ between $X$ and $Y$, and an intermediate variable $Z$ in $p$, we denote by $p[X..Z]$ the path consisting of the edges of $p$ that appear between $X$ and $Z$.

Variable $Z$ is a *collider* in path a $p$ between $X$ and $Y$, if both $p[X..Z]$ and $p[Z..Y]$ point to $Z$. A path that does not have any collider is said to be *unblocked*.

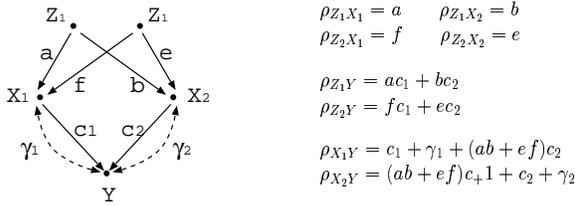

Figure 3: Wright's equations.

The *depth* of a node $Y$ in a causal diagram is defined as the length (i.e., number of edges) of the longest chain from any ancestor of $Y$ to $Y$. Nodes with no ancestors have depth 0.

The next lemma gives a restriction on the depth of intermediate variables in unblocked paths:

**Lemma 1** *Let $p$ be an unblocked path between $X$ and $Y$, and let $Z$ be an intermediate variable in $p$. Then, $depth(Z) < max\{depth(X), depth(Y)\}$.*

## 2.3 Wright's method of Path Analysis

The method of path analysis [13] for identification is based on a decomposition of the correlations between observed variables into polynomials on the parameters of the model. More precisely, for variables $X$ and $Y$ in a recursive model, the correlation coefficient of $X$ and $Y$, denoted by $\rho_{XY}$, can be expressed as:

$$\rho_{X,Y} = \sum_{\text{paths } p_l} T(p_l) \qquad (2)$$

where the term $T(p_l)$ represents the product of the parameters of the edges along path $p_l$, and the summation ranges over all unblocked paths between $X$ and $Y$. Figure 3 shows a simple model and the decompositions of the correlations of each pair of variables.

The set of equations obtained from Wright's decompositions summarizes all the statistical information encoded in the model. Therefore, any question about identification can be decided by studying the solutions for this system of equations.

## 3 Analysis of Identification

The starting point for our analysis is the set of equations provided by Wright's decomposition of correlations. Each term in this decomposition corresponds to an unblocked path in the causal diagram. Now, observe that if we have two edges pointing to the same variable, say $Y$, then they cannot both appear in an unblocked path (because otherwise $Y$ would be a collider blocking the path). Hence, the expressions for the decomposition of correlations are linear on the parameters of any subset of edges incoming a variable $Y$ (i.e., edges with an arrow head pointing to $Y$). This observation leads to the following method to decide the identification of the model.

First, partition all the edges in the causal diagram into subsets of incoming edges. Then, study the identification of the parameters associated with each subset by analyzing the solution of a system of linear equations.

Two conditions must be satisfied to obtain the identification of the parameters. First, there must exist a sufficient number of linearly independent equations. Second, the coefficients of these equations, which are functions of other parameters in the model, must be identified.

To address the first issue, we obtained a graphical characterization for linear independence, called the G Criterion. The second point is addressed by establishing an appropriate order to solve the systems of equations.

The following sections will formally develop this graphical analysis of identification.

### 3.1 Basic Systems of Linear Equations

We begin by partitioning the set of edges in the causal diagram into subsets of incoming edges.

Fix an ordering $\Delta$ for the variables in the model, with the only restriction that if $depth(X) < depth(Y)$, then $X$ must appear before $Y$ in $\Delta$. For each variable $Y$, we define $Inc(Y)$ as the set of edges in the causal diagram that connect $Y$ to any variable appearing before $Y$ in the ordering $\Delta$.

The next lemma formalizes some observations made above.

**Lemma 2** *Any unblocked path between $Y$ and some variable $Z$ can include at most one edge from $Inc(Y)$. Moreover, if $depth(Z) \leq depth(Y)$, then any such path must include exactly one edge from $Inc(Y)$.*

Now, fix an arbitrary variable $Y$, and let

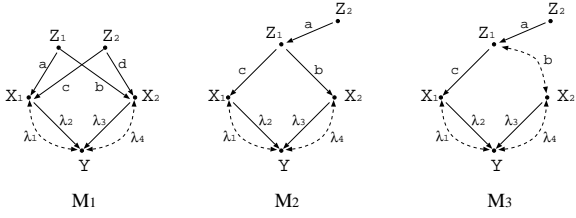

Figure 4: Models $M_1$, $M_2$ and $M_3$

$\lambda_1, \ldots, \lambda_m$ denote the parameters of the edges in $Inc(Y)$. Then, Lemma 2 allows us to express the correlation between $Z$ and $Y$ as a linear equation on the $\lambda_j$'s:

$$\rho_{Z,Y} = a_0 + \sum_{j=1}^{m} a_j \cdot \lambda_j$$

Observe that the independent term $a_0$ is 0 if $depth(Z) \leq depth(Y)$.

Now, given a set of variables $\mathbf{Z} = \{Z_1, \ldots, Z_k\}$, we let $\Phi_{\mathbf{Z},Y}$ [1] denote the system of equations corresponding to the decompositions of correlations $\rho_{Z_1Y}, \ldots, \rho_{Z_kY}$:

$$\Phi_{\mathbf{Z},Y} = \begin{cases} \rho_{Z_1Y} = a_{10} + \sum_{j=1}^{m} a_{1j} \cdot \lambda_j \\ \ldots \\ \rho_{Z_kY} = a_{k0} + \sum_{j=1}^{m} a_{kj} \cdot \lambda_j \end{cases} \quad (3)$$

### 3.2 Auxiliary Sets and Linear Independence

Following the ideas presented in the beginning of the section, we want to find a set of variables that provides a system of linearly independent equations. This motivates the following definition:

**Definition 3 (Auxiliary Set)** *A set of variables $\mathbf{Z} = \{Z_1, \ldots, Z_k\}$ is an Auxiliary Set for $Y$ if $|\mathbf{Z}| = |Inc(Y)|$ and the system of equations $\Phi_{\mathbf{Z},Y}$ is linearly independent.*

Next, we obtain a graphical characterization for Auxiliary Sets. We first analyze a few examples, and then introduce our G criterion.

---

[1] Whenever clear from the context, we drop the reference to $Y$ and simply write $\Phi_{\mathbf{Z}}$.

Consider the models in Figure 4. In each of those cases, the only possible choice for an auxiliary set for $Y$ is $\{X_1, X_2, Z_1, Z_2\}$. However, this set only satisfies the definition for models $M_1$ and $M_3$. The problem with model $M_2$ involves variables $Z_1$ and $Z_2$, because the decomposition of their correlation with $Y$ are not linearly independent:

$$\begin{cases} \rho_{Z_1Y} = a\lambda_2 + b\lambda_3 \\ \rho_{Z_2Y} = ca\lambda_2 + cb\lambda_3 = c \cdot [a\lambda_2 + b\lambda_3] \end{cases}$$

This situation is reflected in the causal diagram by the fact that every unblocked path between $Z_1$ and $Y$ can be extended by the edge $Z_2 \to Z_1$ to give an unblocked path between $Z_2$ and $Y$, and those are all unblocked paths between $Z_2$ and $Y$.

The problem is avoided in the other models because, in $M_1$ there exist disjoint paths connecting $Z_1$ and $Z_2$ to $Y$, and in $M_3$ if we extend the path $Z_1 \leftrightarrow X_2 \to Y$ with the edge $Z_2 \to Z_1$ we obtain a blocked path.

In general, the situation can become much more complicated, with one equation being a linear combination of several others. However, these examples illustrate the essential graphical properties that characterize linear independence.

**G Criterion:** *A set of variables $\mathbf{Z} = \{Z_1, \ldots, Z_k\}$ satisfies the G criterion with respect to $Y$ if there exist $p_1, \ldots, p_k$ such that:*

(i) *$p_i$ is an unblocked path between $Z_i$ and $Y$ including some edge from $Inc(Y)$;*

(ii) *If paths $p_i$ and $p_j$ have a common variable $U$, then either*

  a) *both $p_i[Z_i..U]$ and $p_j[U..Y]$ point to $U$; or*

  b) *both $p_j[Z_j..U]$ and $p_i[U..Y]$ point to $U$.*

Note that the second condition above basically states that two paths $p_i$ and $p_j$ cannot be broken at a common variable $U$ and their pieces be rearranged to form two unblocked paths.

As it turns out, the graphical conditions in the G criterion precisely characterize the linear independence of the system (3). This is formally stated in the next theorem (see appendix A for a proof):

**Theorem 1** *A set of variables $\mathbf{Z} = \{Z_1, \ldots, Z_k\}$, with $|\mathbf{Z}| = |Inc(Y)|$, is an auxiliary set for $Y$ if and only if it satisfies the G criterion.*

### 3.3 Model Identification Using Auxiliary Sets

Suppose now that we can find an auxiliary set $\mathcal{A}_Y$ for each variable $Y$ in the model. This implies that for each $Y$ there exists a system of linear equations $\Phi_{\mathcal{A}_Y}$ that can be solved uniquely for the parameters $\lambda_1, \ldots, \lambda_m$ of the edges in $Inc(Y)$. This fact, however, does not guarantee the identification of the $\lambda_i$'s, because the solution for each $\lambda_i$ is a function of the coefficients in the linear equations, which may depend on non-identified parameters.

To prove identification we need to find an appropriate order to solve the systems of equations. Let us consider a simple situation. Suppose that for each variable $Y$ the following condition holds:

$$depth(Z_i) < depth(Y), \text{for all } Z_i \in \mathcal{A}_Y \quad (4)$$

Now, consider the linear equation provided by the decomposition of $\rho_{Z_iY}$. The coefficients in this equation are sums of terms associated with unblocked paths between $Z_i$ and $Y$. From condition (4) and lemma 1, it follows that all such paths include only variables at depth smaller than $depth(Y)$. Thus, if we solve the systems associated with all those variables before solving $\Phi_{\mathcal{A}_Y}$, then the coefficients of $\Phi_{\mathcal{A}_Y}$ will be identified.

**Theorem 2** *If every variable $Y$ has an auxiliary set satisfying condition (4), then the model is identified.*

In the general case, however, the auxiliary set for some variable $Y$ may contain variables at greater depths than $Y$, or even descendants of $Y$. This forces us to solve the systems of equations in a different order than the one established by the depth of the variables.

A close inspection on the coefficients of $\Phi_{\mathcal{A}_Y}$ shows that it is sufficient to solve the systems associated with some $Z_i$'s in $\mathcal{A}_Y$ before solving $\Phi_{\mathcal{A}_Y}$. There are basically two cases:

1) $Z_i$ is a descendant of $Y$; or

2) $Z_i$ is a non-descendant of $Y$, but there is an unblocked path between $Z_i$ and $Y$ of the form $Z_i \leftarrow \ldots \leftarrow \cdot \leftrightarrow Y$.

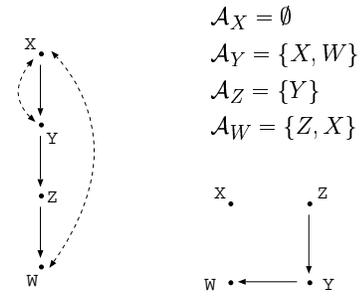

Figure 5: Example Auxiliary Sets method.

We can represent those restrictions by a directed graph, called dependence graph.

The next theorem states our general sufficient condition for model identification (see appendix A for a proof):

**Theorem 3** *If there exist auxiliary sets for the variables in the model such that the associated dependence graph is acyclic, then the model is identified.*

Figure 5 shows an example that illustrates the method just described. Apparently, this is a very simple model. However, it actually requires the full generality of our method.

## 4 Discussion

The graphical condition presented in this paper is the most general sufficient condition for identification of recursive SEM available in the literature. Hence, a natural question is whether it is also necessary for the identification of the model.

We first observe that it is not hard to derive a proof for the non-identification of the model if there exists a variable with no auxiliary set. Thus, we only need to investigate if models with a cyclic dependence graph are non-identified.

An interesting situation occurs if the dependence graph has a single cycle. For example, suppose there is a cycle with the variables $X$, $Y$ and $Z$ (i.e., $X \to Y \to Z \to X$). If this is the only cycle in the graph, then we may solve all systems that need to be solved before $\Phi_{\mathcal{A}_X}$, except for $\Phi_{\mathcal{A}_Z}$.

At this point, we fix some parameter in $\Phi_{\mathcal{A}_X}$, say $\lambda$, as a constant, and remove the equation associated with $\rho_{ZX}$ from $\Phi_{\mathcal{A}_X}$. Using the remaining equations in $\Phi_{\mathcal{A}_X}$, we obtain expres-sions for the other parameters in terms of $\lambda$. Once $\Phi_{\mathcal{A}_X}$ is solved, we can proceed to solve $\Phi_{\mathcal{A}_Y}$ and then $\Phi_{\mathcal{A}_Z}$, obtaining expressions in terms of $\lambda$ for all parameters in those systems. Finally, substituting those expressions back into the equation associated with $\rho_{ZX}$, we obtain a polynomial on the parameter $\lambda$. If the polynomial does not vanish, this implies that $\lambda$ can assume only a finite number of distinct values (namely, the roots of the polynomial). For each such value we have a distinct parameterization for the model that generates the same covariance matrix. Hence, the model is k-identified, for some k. In [3] we present a 2-identified model.

We believe that the polynomial mentioned above never vanishes, and conjecture that if the dependence graph has only isolated cycles (i.e., cycles with no common variable) then the model is k-identified.

When the dependence graph has multiple cycles with common variables, the application of the method above leads to systems of nonlinear equations on two or more parameters. Perhaps a closer examination of the structure of these systems may allow to decide the identification status of the model.

In [4], we provided a procedure to find an auxiliary set for a given variable $Y$. The procedure reduces the problem to a max-flow computation and executes in time $O(n^3)$. Together with Theorem 2, this gives an algorithm for testing the identification status of the model. An algorithm implementing the more general result of Theorem 3 would require finding sets of auxiliary variables that give rise to an acyclic dependence graph.

## Appendix A

Due to space constraints, we only sketch the proofs of theorems 1 and 3, and refer the reader to [3] for full proofs.

**Proof of Theorem 1:**

The system of equations $\Phi_{\mathbf{Z}}$ can be written in matrix form as:

$$\rho = A \cdot \Lambda$$

where $\rho = [(\rho_{Z_1 Y} - a_{10}) \ldots (\rho_{Z_k Y} - a_{k0})]'$, $A = [a_{ij}]$ is a $k$ by $k$ matrix, and $\Lambda = [\lambda_1 \ldots \lambda_k]'$.

We prove the theorem by analyzing the determinant of $A$, which is given by

$$Det(A) = \sum_{\sigma} (-1)^{|\sigma|} \prod_{j=1}^{k} a_{j\sigma(j)} \qquad (5)$$

where the summation ranges over all permutations of $\langle 1, \ldots, k \rangle$, and $|\sigma|$ denotes the parity of permutation $\sigma$.

First, suppose that $\mathbf{Z} = \{Z_1, ..., Z_k\}$ satisfies the G criterion with respect to $Y$, and let $p_1, ..., p_k$ witness this fact. Without loss of generality, assume that path $p_i$ connects variable $Z_i$ to $Y$ and includes the edge from $Inc(Y)$ with parameter $\lambda_i$. Observe that entry $a_{ij}$ of $A$ is given by a sum of terms associated with paths between $Z_i$ and $Y$ that include the edge with parameter $\lambda_j$. Thus, we can write $a_{ii} = \left(\frac{T(p_i)}{\lambda_i} + a'_{ii}\right)$, and this shows that the term $\tau = \left[\prod_i \frac{T(p_i)}{\lambda_i}\right]$ appears in the summation on the right-hand side of 5.

In fact, every term in the summation of 5 is given by the product of terms associated with unblocked paths between each of the $Z_i$'s and $Y$. However, it follows from condition $(ii)$ of the G criterion that the edges that compose $p_1, ..., p_k$ cannot be rearranged to form a distinct set of paths $p'_i, ..., p'_k$ connecting the $Z_i$'s to $Y$ (The proof of this fact is somewhat technical, but not too difficult). In particular, this implies that $\tau$ is not cancelled out by any other term in (5). Hence, the determinant of $A$ is a polynomial that is not identically zero, and vanishes on a set of Lebesgue measure zero.

The converse is proved by observing that, if condition $(ii)$ of the G criterion does not hold, then for any set of unblocked paths $\{p_1, \ldots, p_k\}$ connecting the $Z_i$'s to $Y$, there exists a pair, say $p_i$ and $p_j$, with a common variable $U$ such that the paths formed by the concatenations

$$p'_i = p_i[Z_i..U] \cdot p_j[U..Y]$$
$$p'_j = p_j[Z_i..U] \cdot p_i[U..Y]$$

are unblocked. Now, the term associated with $\{p_1, \ldots, p_k\}$ in (5) is the same as the one associated with $\{p_1, \ldots, p'_i, p'_j, \ldots, p_k\}$, but they appear in permutations with opposite parities, and hence are cancelled out. This argument is extended to the general case, with multiple intersections, by an inductive argument. □

**Proof of Theorem 3:**

The systems of equations are solved according to the partial order defined by the dependence graph. We prove the theorem by showing that at the time of solving $\Phi_{\mathcal{A}_Y}$ every coefficient in this system is identified.

Fix a variable $Y$ in the model, and let $Z \in \mathcal{A}_Y$. Next, we examine the coefficients in the decomposition of $\rho ZY$. There are three cases:

Case 1: $Z$ is non-descendant of $Y$, and there is no unblocked path between $Z$ and $Y$ of the form $Z \leftarrow \ldots \leftrightarrow Y$.

The decomposition of $\rho_{ZY}$ can be written as

$$\rho_{ZY} = \sum_i c_i \delta_i + \sum_j b_j \lambda_j$$

where the $\delta_i$'s are the parameters of the directed edges in $Inc(Y)$ (e.g., $X_i \to Y$), and the $\lambda_j$'s are the parameters of bidirected edges in $Inc(Y)$ (e.g., $V_j \leftrightarrow Y$). Then the coefficients in 4 are given by

- $c_i = \rho_{ZX_i}$ and $b_j = \begin{cases} 1, & \text{if } Z = V_j \\ 0, & \text{otherwise} \end{cases}$

This follows because

1) The set of unblocked paths between $Z$ and $Y$ that include $(X_i \to Y)$ is precisely the set of all unblocked paths between $Z$ and $X_i$ extended by $(X_i \to Y)$. Thus, $c_i = \rho_{ZX_i}$.

2) If $Z = V_j$, for some $j$, then there is only one unblocked path between $Z_i$ and $Y$ including $(V_j \leftrightarrow Y)$, which is composed by this single edge. Thus, $b_j = 1$ in this case.

3) Otherwise, observe that any unblocked path between $Z$ and $Y$ including $(V_j \leftrightarrow Y)$ has the form $Z \leftarrow \ldots V_j \leftrightarrow Y$. Since we assume no such paths exists, we have $b_j = 0$.

Cases 2 and 3 correspond to the situation where there is some unblocked path between $Z$ and $Y$ that ends with a bidirected edge, or $Z$ is a descendant of $Y$. We ommit the proof of those cases due to space constraints.

□